\documentclass[11pt]{article}
\usepackage[utf8]{inputenc}
\usepackage[T1]{fontenc}
\usepackage{graphicx}
\usepackage{longtable}
\usepackage{wrapfig}
\usepackage{rotating}
\usepackage[normalem]{ulem}
\usepackage{capt-of}
\usepackage{hyperref}
\usepackage{mmt}
\author[1,2]{Michael T.M.B. Morris-Thomas\thanks{E-mail: michael.morris-thomas@me.com}}
\author[2]{Marius Martens}
\affil[1]{Oceans Graduate School, The University of Western Australia, Crawley, Western Australia}
\affil[2]{Worley Offshore Energy, Perth, Western Australia}
\setlist{itemsep=2pt, parsep=0pt, topsep=6pt, partopsep=0pt}
\date{\today}
\title{An application of machine learning to the motion response prediction of floating assets}
\hypersetup{
             pdfauthor={Michael T.M.B. Morris-Thomas},
             pdftitle={An application of machine learning to the motion response prediction of floating assets},
             pdfkeywords={gradient boosting machine, FPSO, Machine Learning, metocean, mooring system, multivariate regression, weathervaning},
             pdfsubject={Uses various multivariate regression models to predict response variables associated with a single point mooring system},
             pdfcreator={Emacs 30.1 (Org mode 9.7.11)}, 
             pdflang={English},
             colorlinks=true,
             linkcolor=RoyalBlue,
             citecolor=BrickRed,
             urlcolor=RoyalBlue}
\usepackage{biblatex}
\begin{document}

\maketitle
\begin{abstract}
The real-time prediction of floating offshore asset behavior under
stochastic metocean conditions remains a significant challenge in
offshore engineering. While traditional empirical and frequency-domain
methods work well in benign conditions, they struggle with both
extreme sea states and nonlinear responses. This study presents a
supervised machine learning approach using multivariate regression to
predict the nonlinear motion response of a turret-moored vessel in
400 m water depth. We developed a machine learning workflow combining
a gradient-boosted ensemble method with a custom passive weathervaning
solver, trained on approximately \(10^6\) samples spanning
100 features. The model achieved mean prediction errors of less than
5\% for critical mooring parameters and vessel heading accuracy to
within 2.5 degrees across diverse metocean conditions, significantly
outperforming traditional frequency-domain methods. The framework has
been successfully deployed on an operational facility, demonstrating
its efficacy for real-time vessel monitoring and operational
decision-making in offshore environments.
\end{abstract}
\section{Introduction}
\label{sec:orgf5af340}

\begin{figure}[t]
\centering
\includegraphics[scale=1.0]{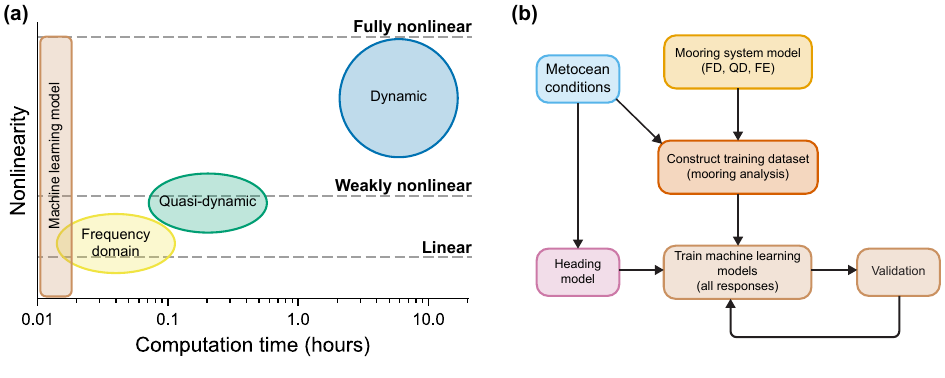}
\caption{\label{fig:computation-workflow}Mooring assessment approaches and a machine learning workflow. (a) Nonlinearity versus indicative computation time for a typical mooring analysis: frequency domain (FD), \(<10\) seconds; quasi dynamic (QD), order 10 minutes (single realization); and a fully "dynamic" finite element representation of the mooring system (FE), order \(6\text{--}12\) hours (single realization). (b) Proposed machine learning workflow for model training on vessel and mooring system responses.}
\end{figure}

Of interest to offshore engineers, oceanographers, and naval
architects is the ability to accurately forecast and predict in
real-time the behavior of offshore floating assets under the influence
of stochastic metocean conditions. Historically, this challenge has
been addressed through a combination of empirical and frequency-domain
methods. While these approaches are effective in benign metocean
conditions for a well-behaved floating system, they struggle in
extreme sea states or when the system's motion response is
nonlinear. Although this can be remedied through nonlinear time-domain
computations over several sea state realizations, such schemes are
typically too computationally intensive and impractical for real-time
application (see Fig. \ref{fig:computation-workflow}a). Recently,
sophisticated and mature machine learning (ML) algorithms have
provided an alternative and more viable approach. Such algorithms are
computationally efficient and, when appropriately trained, have shown
particular promise in capturing the complex nonlinear relationships
inherent in wave-structure interaction problems.

Machine learning has recently enabled solutions to complex and
computationally demanding problems in science and engineering with
impressive results
\autocite{LeCunEtAl2015,BauerEtAl2015,BruntonEtAl2020,RegenwetterEtAl2022,KochkovEtAl2024}.
While its application to offshore engineering began relatively
recently \autocite{GuarizeEtAl2007}, it has gained traction in predicting
vessel motions and mooring system responses to metocean conditions
\autocite{DePinaEtAl2013,AlvarellosEtAl2021,GuoEtAl2022}. Of these,
most approaches focus on time-series prediction using deep learning
frameworks, particularly convolutional neural networks and long
short-term memory models. Although these methods show promise
\autocite{WangEtAl2022,GrafeEtAl2024}, they face limitations in
generalizing across diverse metocean conditions and capturing extreme
response statistics for given sea states.

For offshore facility monitoring and decision-making, real-time
knowledge of these response statistics is crucial when assessing a
facility's performance against its limit state design
criteria. Motivated by a requirement to predict, in real-time, the
performance of a floating facility against its design criteria, we
outline an alternative approach.

In this work, we focus on a generic turret-moored vessel designed to
passively weathervane about its mooring center in approximately 400 m
water depth. This type of facility presents unique challenges due to
the coupling of its heading with incident environmental forces that
subsequently govern both its motion response and mooring system
behavior. Using approximately 495 years of metocean combinations, we
constructed a training dataset through frequency-domain screening
followed by time-domain computations for critical metocean conditions.

Our ML workflow combines multivariate regression with a
custom-developed passive weathervaning solver. This physics-based
vessel heading prediction outperforms ML methods in both accuracy and
computational efficiency. To predict system response statistics, we
focus on: algorithm selection, feature importance, and dimensional
reduction. We illustrate the validation of our machine learning
pipeline, which combines an XGBoost \autocite{ChenGuestrin2016}
gradient-boosted ensemble method with a custom-developed passive
weathervaning solver. This pipeline was trained on a moderately sized
metocean and mooring response dataset comprising \(N\approx10^6\)
samples with around 100 unique features that encompass waves, wind,
and current, and various parameters characterizing the system's
mooring behavior and hydrodynamic performance.

We demonstrate the model's utility in predicting key mooring
responses, including vessel heading, turret excursion and direction,
and both fairlead and anchor tensions. The results show a marked
improvement in prediction accuracy when compared to traditional
frequency-domain methods, with mean prediction errors of approximately
5\% for critical parameters and the vessel's heading resolved to
within an absolute error of 2.5 degrees. Notably, the model maintains
this level of performance across a wide range of metocean conditions
comprising combinations of current, wind, and bi-modal sea states.

Finally, we briefly discuss the applicability of our scheme for
real-time prediction and forecasting purposes. While this study
focuses on a generic turret-moored system to demonstrate the model and
its performance, the developed workflow has been successfully deployed
on an operational facility and is currently providing real-time
predictions in a production environment to support turret
disconnection decisions. This practical validation demonstrates that
our framework can be effectively integrated with
existing vessel monitoring systems, providing asset owners with a
robust solution for operational decision-making in offshore
environments.

The paper is organized as follows. Section \ref{sec:design-data}
presents our input data, including: the underlying metocean dataset
(Sec. \ref{sec:metocean-data}), the turret-moored vessel and associated
numerical models (Sec. \ref{sec:vessel-modeling}), and the construction
of the mooring response dataset and critical system responses
(Sec. \ref{sec:mooring-response-dataset}). Section \ref{sec:ml-workflow}
covers our machine learning workflow
(Fig. \ref{fig:computation-workflow}b), including: heading analysis
(Sec. \ref{sec:heading-analysis}), regression models
(Sec. \ref{sec:regression-models}), baseline training
(Sec. \ref{sec:baseline-training}), feature selection, and
hyperparameter tuning (Sec. \ref{sec:hyperparameter-tuning}) to
facilitate model selection. We then present the results and validation
of our workflow in Sec. \ref{sec:results} along with a brief discussion
on implementation (Sec. \ref{sec:practical-application}). Finally,
Sec. \ref{sec:conclusion} provides conclusions and recommendations for
further work.
\section{Design data}
\label{sec:design-data}
\subsection{Metocean data}
\label{sec:metocean-data}
The metocean dataset was compiled from three primary sources: an
extensive hindcast dataset, synthetically generated conditions derived
from kernel density estimation (KDE) of the hindcast data, and
synthetic tropical low conditions including a small subset of hindcast
tropical cyclone conditions. For this study, we utilized a subset of a
full production dataset. It comprised 25.5 million unique data points
and represents 495 years of 3-hour sea states (1.5 million metocean
conditions). Summary statistics for the metocean dataset are provided
in Tab. \ref{tab:metocean-data}.

The environmental parameters describing each unique metocean
combination included two distinct wave systems, a wind component, and
a single current component. The first wave system represented
wind-generated waves with relatively low periods, while the second
characterized swell conditions with longer periods. The energy content
of both wave systems was individually described by 5-parameter
JONSWAP spectra \autocite{HasselmannEtAl1973} with spectral bandwidth and
peak shape factors defined according to Lewis and Allos'
parameterization \autocite{LewisAllos1990}. Wave steepness limits were
enforced in accordance with Det Norske Veritas (DNV) guidelines
\autocite{DNV-RP-C205_2021}. Wind conditions were specified by direction
and speed with energy content following the NPD spectrum model
\autocite{ISO2015}, and current was similarly defined by direction and
speed but modeled as a constant free-surface velocity.

\begin{table*}[t]
\caption{\label{tab:metocean-data}Summary statistics for the metocean dataset. All environmental directions are defined as "from," see Fig. \ref{fig:coordinates}. Wave parameters: \(H_s\) -- significant wave height; \(T_p\) -- peak spectral wave period; \(\theta_p\) -- peak spectral wave direction. Wind parameters: \(U_w\) -- wind speed; \(\theta_w\) -- wind direction. Current parameters: \(U_c\) -- surface current speed; \(\theta_c\) -- current direction. Statistical measures: \(\mu\) -- mean; \(\sigma\) -- standard deviation; \(Q_{25}, Q_{50}, Q_{75}\) -- 25th, 50th, and 75th percentiles.}
\centering
\begin{tabular*}{\textwidth}{@{\hspace{6pt}}l@{\extracolsep{\fill}}lllllllll@{\hspace{6pt}}}
\toprule
Environment & Parameter & Unit & \(\mu\) & \(\sigma\) & min & \(Q_{25}\) & \(Q_{50}\) & \(Q_{75}\) & max\\
\midrule
\textbf{Wave \#1} & \(H_{s,1}\) & m & 2.60 & 2.05 & 0.26 & 1.24 & 1.75 & 2.72 & 8.00\\
 & \(T_{p,1}\) & s & 8.44 & 1.87 & 2.85 & 7.52 & 8.49 & 8.94 & 17.00\\
 & \(\theta_{p,1}\) & degrees & 200.49 & 80.26 & 0.00 & 209.50 & 223.14 & 237.25 & 359.93\\
\midrule
\textbf{Wave \#2} & \(H_{s,2}\) & m & 1.79 & 0.72 & 0.00 & 1.34 & 1.82 & 2.03 & 8.95\\
 & \(T_{p,2}\) & s & 13.08 & 2.87 & 0.00 & 11.63 & 13.22 & 15.00 & 40.00\\
 & \(\theta_{p,2}\) & degrees & 223.61 & 34.46 & 0.00 & 223.25 & 225.00 & 232.74 & 359.89\\
\midrule
\textbf{Wind} & \(U_w\) & m/s & 9.93 & 7.73 & 0.00 & 5.02 & 7.48 & 11.10 & 41.76\\
 & \(\theta_w\) & degrees & 178.83 & 84.62 & 0.00 & 120.00 & 192.37 & 233.87 & 359.98\\
\midrule
\textbf{Current} & \(U_c\) & m/s & 0.51 & 0.56 & 0.00 & 0.14 & 0.29 & 0.64 & 4.65\\
 & \(\theta_c\) & degrees & 137.94 & 99.86 & 0.00 & 57.30 & 110.70 & 213.84 & 360.00\\
\bottomrule
\end{tabular*}
\end{table*}
\subsection{Vessel and mooring system modeling}
\label{sec:vessel-modeling}
\begin{figure}[t]
\centering
\includegraphics[scale=1.0]{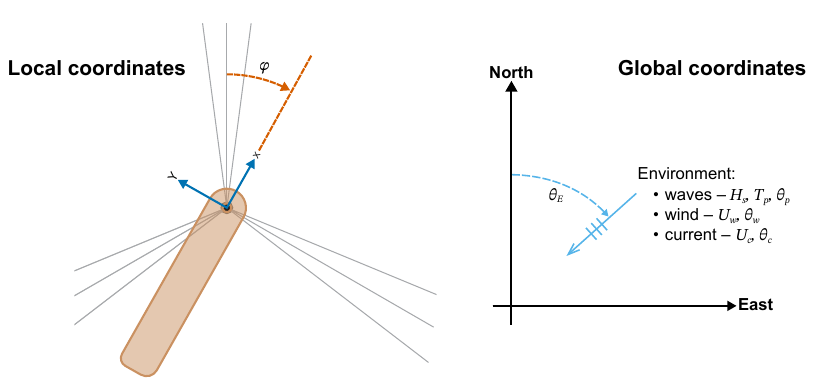}
\caption{\label{fig:coordinates}Local and global coordinate systems for a single point mooring (SPM) system. The vessel heading in global coordinates is depicted by \(\varphi\) and \(\theta_{E}\) denotes an environment direction (wave, wind, or current). Mooring lines (light gray) are illustrated in a \(3 \times 3\) arrangement centered about the turret (black circle).}
\end{figure}

We adopted a generic weathervaning Floating Production, Storage, and
Offloading (FPSO) facility (see Fig. \ref{fig:coordinates}) for this
study with its principal particulars characteristic of a Very Large
Crude Carrier (VLCC) class hull form --- length overall \(335\m\), beam
\(58\m\), and moulded depth \(30\m\). The vessel is catenary-moored via an
internal turret in a \(3 \times 3\) single point mooring (SPM)
arrangement about its geostationary center in water of depth
approximately \(400\m\). To facilitate analysis, a numerical model
comprising the vessel and its riser, turret, and mooring system was
constructed in OrcaFlex \autocite{orcina2024a}.

The vessel's hydrodynamic properties were calculated over several
loading conditions through a radiation/diffraction analysis conducted
in OrcaWave \autocite{orcina2024}. Moreover, this provided wave-frequency
transfer functions and low-frequency drift force transfer functions
for the submerged hull. All vessel response transfer functions were resolved for 100
circular wave frequencies discretized over the range \(\omega=0.05
\text{--}2.0\radps\) and encounter angles ranging from \(\theta=0\degree
\text{--} 360\degree\) at 10\(\degree\) intervals. Irregular frequencies
were removed by supplying an internal free-surface mesh
\autocite{orcina2024} to the boundary element method solver. Given the
water depth, the Quadratic Transfer Functions, describing the
low-frequency vessel response, were resolved via Newman's
\autocite{Newman1974a} approximation and the corresponding wave-drift
damping associated with both slow drift and wave-current interaction
accounted for via Aranha's \autocite{Aranha1994} approximation.

In terms of system damping, the quadratic (viscous) roll damping was
numerically computed \autocite{ittc2021} for each vessel loading
condition. Its nonlinear contribution was applied directly during all
time-domain simulations. However, for all frequency-domain simulations
it was linearized using a probabilistic equivalent damping under the
assumption of a Gaussian roll response for each unique sea state in
the metocean dataset.

The mooring and riser system damping was decomposed into two principal
components: a linear contribution in phase with vessel velocity; and a
nonlinear contribution proportional to the mooring and riser system's
cross-flow drag. Considering both forced-oscillations and plane-flow,
both components were computed numerically from a complete Finite
Element representation of the mooring and riser system.

\subsection{Mooring response dataset}
\label{sec:mooring-response-dataset}
Rather than conducting full time-domain simulations over multiple
realizations per metocean combination, we adopted a more
computationally efficient approach to construct the mooring response
dataset. We decomposed the turret-moored vessel model into three
distinct numerical models (see Fig. \ref{fig:computation-workflow}b):

\begin{enumerate}
\item Frequency-domain (FD) model --- an efficient simplified model
decomposed into low-frequency and wave-frequency solution spaces to
screen the complete metocean dataset (Tab. \ref{tab:metocean-data})
\item Quasi-dynamic (QD) model --- a time-domain model using canonical analytic
catenaries \autocite{orcina2024a} with additional damping to
account for cross-flow drag from the mooring lines and risers.
\item Full dynamic (FE) model --- employs a complete finite element
representation of the mooring and riser system. Given its runtime
(approximately \(10~\mathrm{hours}\) per realization), it was applied selectively to:
\begin{itemize}
\item quantify dynamic amplification from mooring line inertia;
\item quantify mooring and riser damping for the QD and FD models.
\end{itemize}
\end{enumerate}

The FD screening identified metocean combinations that produced
mooring system and vessel responses near the system's capacity
limits. For these cases, we replaced the FD results with corresponding
QD model computations.  Each QD simulation comprised ten 3-hour
realizations with a 30-minute ramp time per metocean
combination. Response statistics were calculated through a Gumbel
extreme value analysis.

Critical response statistics of interest included:

\begin{itemize}
\item mean vessel heading and environment encounter angles (from True North)
\item most probable maximum (MPM) turret offset from the mooring center and its corresponding direction
\item MPM mooring line fairlead tension
\item MPM mooring line anchor tension
\item wave-frequency and low-frequency vessel motions.
\end{itemize}

We augmented all mooring system responses from the FD and QD models
with dynamic amplification factors from the FE model. The combined
dataset from this workflow produced the training set.
\section{Machine learning workflow}
\label{sec:ml-workflow}
In reference to Fig. \ref{fig:computation-workflow}b, our methodology comprised a
systematic approach to constructing an ML model capable of predicting
response statistics for a turret-moored system
(Fig. \ref{fig:coordinates}) for a variety of metocean
conditions.
\subsection{Heading analysis}
\label{sec:heading-analysis}
A turret-moored vessel weathervanes about its turret center to find an
equilibrium heading that minimizes the combined environmental forces
from waves, wind, and current. This equilibrium heading can be
determined through a quasi-static analysis by treating the net restoring
moment \(M\) about the turret as a conservative force field. The
corresponding potential energy as a function of vessel heading
\(\varphi\) is:
\begin{equation}
\label{eq:PE}
V(\varphi) = - \int M(\varphi) \, d\varphi,
\end{equation}
\noindent which is valid over the domain
\(\varphi\in[0\degree,360\degree]\). This can be minimized numerically
using standard optimization techniques:
\begin{equation}
\label{eq:minimize}
\varphi_{\mathrm{eq}} = \argmin_{\varphi \in [0\degree,360\degree]} V(\varphi)
\end{equation}
\noindent to yield a single stable global equilibrium vessel heading
\(\varphi_{\mathrm{eq}}\) for a given metocean combination.

The minimization of potential energy may yield multiple local minima,
each representing a stable equilibrium heading, when the environmental
forcing moments are of comparable magnitude. This occurs particularly
when two or more of the wave, wind, and current moments have similar
orders of magnitude. While the global minimum represents the most
energetically favorable heading, these additional stable
configurations can be physically realized depending on the vessel's
initial conditions and dynamic path. However, both these dynamic
effects and secondary equilibria are of lesser importance and
neglected here, focusing solely on the most stable quasi-static
heading.

The computed heading, Eq. \eqref{eq:minimize}, for each metocean
combination was adopted as an additional feature of the training
dataset (see Fig. \ref{fig:computation-workflow}b).
\subsection{Regression models}
\label{sec:regression-models}
Any stationary response statistic \(Y\) of a system with independent
variables \(\mathbf{X} \in \mathbb{R}^{p}\) can generally be expressed
as:
\begin{equation}
\label{eq:model}
Y = f(\mathbf{X}) + \varepsilon
\end{equation}
\noindent where \(f\) denotes an unknown mapping, and \(\varepsilon\) is its
irreducible error --- more loosely speaking: the residual. The
objective is to estimate \(f\) from a set of training data by minimizing
\(\varepsilon\). With this, we consider the following seven regression
models, ranging from linear to ensemble approaches, and a single
neural network approach:

\begin{itemize}
\item \textbf{Linear Regression} (LR) \autocite{HastieEtAl2009}: baseline parametric
approach
\item \textbf{Decision Tree} (DT) \autocite{HastieEtAl2009}: a weak learner that
recursively partitions the feature space through binary splitting
into \(n\) regions with constant predictions
\item \textbf{Random Forest} (RF) \autocite{Breiman2001,HastieEtAl2009}: an
ensemble method that averages predictions from multiple decision
trees through bootstrap sampling
\item \textbf{Histogram Gradient Boosting} (HGB) \autocite{PedregosaEtAl2011}: gradient
boosting with histogram-based splitting
\item \textbf{LightGBM} (LGBM) \autocite{KeEtAl2017}: histogram based gradient boosting
with one-side sampling and exclusive feature bundling
\item \textbf{XGBoost} (XGB) \autocite{ChenGuestrin2016}: gradient boosting with
regularized tree learning
\item \textbf{Multi-Layer Perceptron} (MLP) \autocite{HastieEtAl2009}: a feed forward neural network with a
number of hidden layers.
\end{itemize}

\noindent Apart from LightGBM and XGBoost, all implementations were adopted from the
scikit-learn \autocite{PedregosaEtAl2011} library.

For regression, all angular responses were decomposed into their
orthogonal components (Eastings and Northings) and treated
independently. This approach ensured that all angular responses were
bounded and with periodicity enforced: \(\theta \in [0\degree,
360\degree]\). After regression, the angular responses were
reconstructed from these orthogonal components. In particular, this
method was applied to calculate the direction of the MPM turret
offset, and although not shown here, the approach is also effective in
recovering the vessel heading.
\subsection{Baseline training and features}
\label{sec:baseline-training}
Each regression model, Sec. \ref{sec:regression-models}, can be
optimized by selecting the most efficacious hyperparameters and
features to describe each unique response statistic of the system.

Prior to hyperparameter optimization, baseline regression models for
each response variable were established using standard parameters
commonly referenced in literature. For tree-based methods (RF,
XGB, and LGBM), these included modest ensemble sizes (100 trees), a
standard learning rate (0.1), and a conservative tree depth of 30. For
the MLP model, two hidden layers were used with a rectified linear unit
activation function. Along with Linear Regression, these estimates
provided a baseline for model behavior, feature importance, and
establishing minimum model performance benchmarks before proceeding
with more extensive parameter tuning.

\begin{figure}[t]
\centering
\includegraphics[scale=1.0]{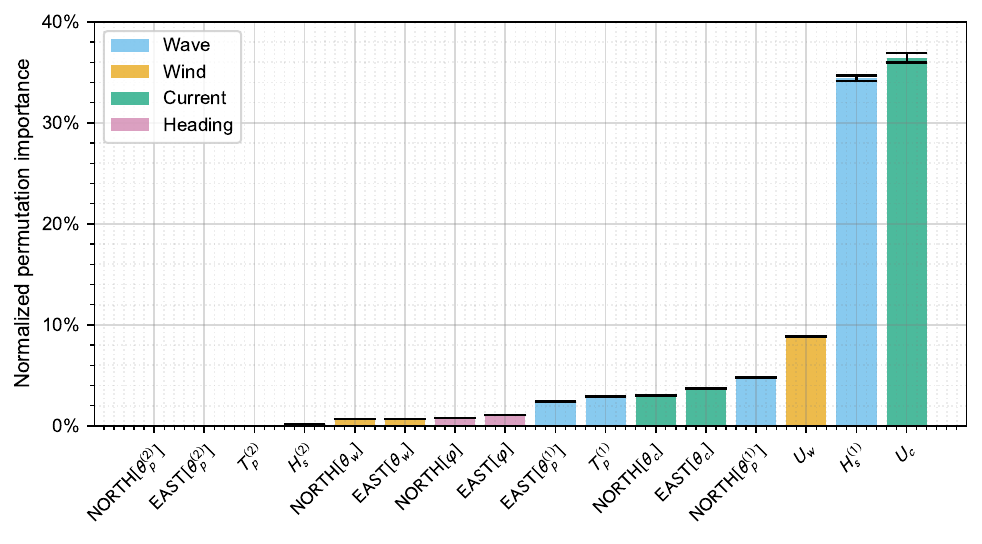}
\caption{\label{fig:perm-imp}Mean permutation importance for the most probable maximum turret offset. East and north components for the environmental directions and vessel heading are denoted \(\mathrm{EAST[\cdot]}\) and \(\mathrm{NORTH[\cdot]}\), respectively. Error bars represent the standard deviation of the mean.}
\end{figure}

To understand the critical metocean parameters, permutation
importance was calculated from baseline estimates from the HGB
model. As an example, this assessment for the MPM turret offset is
illustrated in Fig. \ref{fig:perm-imp}. Apart from vessel heading, which
was evaluated numerically (Sec. \ref{sec:heading-analysis}), the results
indicate that the two most important features, \(U_{c}\) and
\(H_{s}^{(1)}\), dominate over all other parameters with around 35\% --
40\% importance, respectively. Wind emerged as the third most important
parameter with around 9.5\% importance. Presumably, due to a smaller
contributing drift force from much longer spectral peak wave periods,
the secondary wave system appears much further down the permutation
importance list. Of note, is the relative insignificance of the
spectral bandwidth and peak shape factors --- these do not feature in
Fig. \ref{fig:perm-imp} and were consistently shown to be unimportant
across all response statistics. Consequently, these parameters were
excluded from model optimization and final training.

\subsection{Hyperparameter tuning}
\label{sec:hyperparameter-tuning}
With the feature set reduced, we employed an extensive randomized
search hyperparameter optimization strategy \autocite{BergstraBengio2012}
across all critical response variables and regression algorithms
(Sec. \ref{sec:regression-models}). Each critical response variable was
treated independently, and the hyperparameter space of each algorithm
was explored to balance model complexity with computational
feasibility. Throughout, model performance was measured using the Mean
Square Error (MSE) loss function, and high-dimensional hyperparameter
optimization was conducted using a random search cross-validation
strategy \autocite{PedregosaEtAl2011}.

For all tree-based methods (DT, RF, HGB, LGBM, XGB), the optimization
space encompassed both structural parameters and regularization
techniques. In particular, the RF's search space was defined
discretely, exploring ensemble sizes (\(100\text{--}300\) trees), tree depths
(\(10\text{--}30\) levels), and minimum sample constraints for splits and
leaves. For the more sophisticated boosting algorithms (HGB, LGBM,
XGB) continuous distributions were adopted for parameter sampling,
with learning rates logarithmically distributed between \(10^{-3}\) and
\(0.5\), and tree depths ranging from 4 to 20 levels. These models
incorporated additional regularization parameters, including \(L_1\)
(\(\alpha\)) and \(L_2\) (\(\lambda\)) penalties, distributed logarithmically
between \(10^{-8}\) and 10.0, and various sub-sampling ratios for both
observations and features uniformly distributed between 0.6 and 1.0.

For the MLP model, we explored varying network depths and widths,
ranging from simple single-layer configurations (\(50\text{--}200\)
neurons) to deeper architectures of up to four layers with decreasing
widths (200-100-50-25 neurons). The optimization included training
parameters such as learning rate (log-uniform between \(10^{-4}\) and
\(10^{-1}\)), mini-batch sizes (\(32\text{--}256\)), and Adam optimizer
parameters (\(\beta_1\): \(0.8\text{--}0.999\), \(\beta_2\):
\(0.99\text{--}0.999\)).

All regression models incorporated early stopping to prevent
over-fitting, with validation fractions uniformly sampled between 0.1
and 0.2 of the training set, and patience parameters ranging from 10
to 30 epochs to avoid wasted computation on negligible performance
improvement. For reproducibility, all models were initialized with the
same random state.
\section{Results and discussion}
\label{sec:results}
To evaluate model performance, we examined three critical response
statistics: the most probable maximum turret offset from mooring
center, its corresponding offset direction, and the most probable
maximum omni-directional mooring line fairlead tension. Using an
independent validation dataset comprising 20\% of the original metocean
dataset, that was set aside and not utilized during model training, we
assessed each model's ability to predict target values using:

\begin{itemize}
\item Root Mean Squared Error (\(\RMSE\))
\item Mean Absolute Error (\(\MAE\))
\item Coefficient of determination (\(\Rsq\))
\item Residual statistics (mean, minimum, maximum).
\end{itemize}

\noindent The residual for each validation example \(i\) is defined \(e_{i}
= \hat y_{i} - y_{i}\) where \(\hat y_{i}\) and \(y_{i}\) denote the
predicted and actual values, respectively.

This ensemble of metrics provides complementary insights into the
prediction accuracy of each model, with RMSE penalizing larger errors,
MAE indicating average absolute deviation, \(R^{2}\) capturing the
explained variance in the model's prediction, and residual statistics
revealing systematic prediction biases and heteroscedasticity.
\subsection{Linear model}
\label{sec:linear-model}
\begin{figure}[t]
\centering
\includegraphics[scale=1.0]{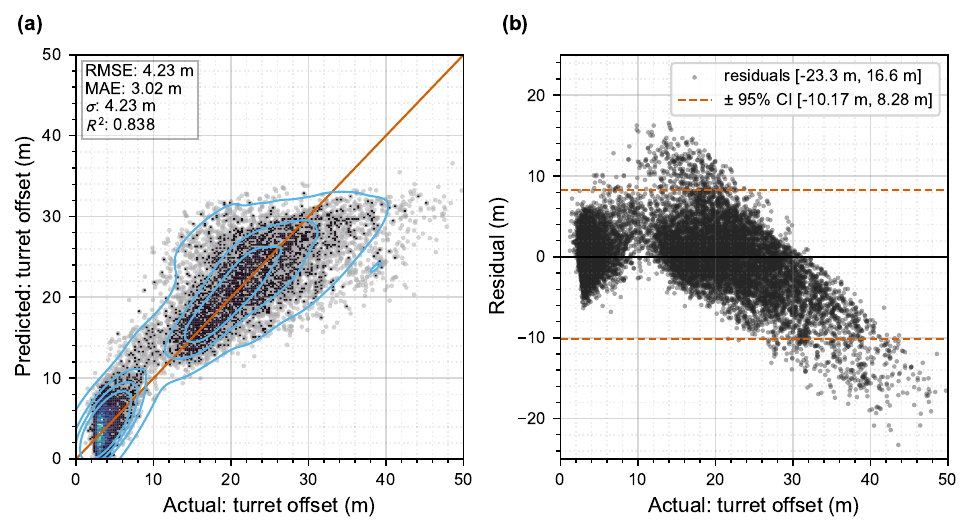}
\caption{\label{fig:turret-offset-linear-QQ-prediction}Linear Regression model performance for the most probable maximum turret offset prediction. (a) Actual versus predicted values with the solid red line denoting a perfect fit and blue contours indicating lines of equal probability density. (b) Model residuals \(e_{i}\) with dashed red lines denoting the 95\% bootstrap confidence intervals (CI) for the mean residual.}
\end{figure}

The baseline linear regression model's performance on the validation
dataset revealed significant limitations in predicting MPM turret
offset, as shown in
Figures \ref{fig:turret-offset-linear-QQ-prediction}a and
\ref{fig:turret-offset-linear-QQ-prediction}b. With an \(\RMSE\) of
\(4.23\m\) and \(\Rsq=0.838\), the model exhibits systematic prediction
errors characterized by a distinct nonlinear pattern in the residual
distribution (Fig. \ref{fig:turret-offset-linear-QQ-prediction}b). Our
analysis revealed a consistent over-prediction in the low-offset
regime (\(<10\m\)) and an under-prediction at higher offsets (\(>30\m\)),
indicating a fundamental mismatch between the linear model assumptions
and the underlying physical relationships. Overall, the residuals span
from \(-23.3\m\) to \(16.6\m\) and demonstrate clear heteroscedasticity,
with the variance increasing proportionally with the predicted
values. While highlighting these systematic biases and the linear
model's limitations, this analysis establishes a quantitative
performance baseline and provides valuable insights into the nonlinear
characteristics of turret offset behavior, thus motivating the
adoption of more sophisticated regression models.
\subsection{Single-shot performance}
\label{sec:single-shot}
\begin{figure}[t]
\centering
\includegraphics[scale=1.0]{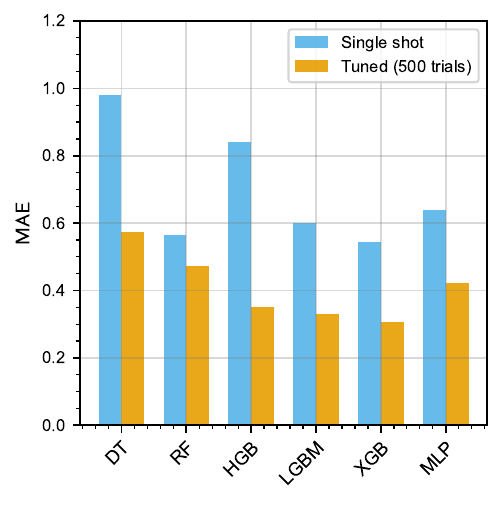}
\caption{\label{fig:single-shot}A Comparison of each model's performance after hyperparameter tuning}
\end{figure}

Using more sophisticated regression models for prediction
(Sec. \ref{sec:regression-models}), Fig. \ref{fig:single-shot} presents a
comparative analysis of the MAE for the turret offset, evaluated both
with default model parameters and after hyperparameter optimization
through 500 trials. In this context, single-shot performance refers to
the model’s baseline ability, without any hyperparameter tuning, to
accurately predict on unseen data that was not explicitly represented
in its training set.

The results demonstrate that systematic hyperparameter tuning
substantially improves model performance across all methods, with
error reductions ranging from 30\% to 60\%. Gradient boosting ensemble
methods (HGB, LGBM, XGB) achieved superior performance after tuning
(\(\MAE \approx 0.3\text{--}0.35\m\)), exhibiting a marked improvement from
their baseline performance values (\(\MAE \approx 0.55\text{--}0.85\m\)). In
contrast, the single decision tree (DT) showed the highest error in
both before and after tuning (\(\MAE=0.98\m\) and \(0.55\m\),
respectively); and, moreover, the RF and MLP approaches demonstrated
intermediate performance (after tuning \(\MAE \approx 0.45\m\)). While
neural networks (MLP model) could theoretically achieve comparable
performance to gradient boosting methods through extensive
hyperparameter optimization, the computational cost and implementation
complexity will likely far outweigh the potential marginal
improvements on offer over a well-tuned gradient boosting model.

\subsection{Model performance and selection}
\label{sec:model-performance}
\begin{figure}[!t]
\centering
\includegraphics[scale=1.0]{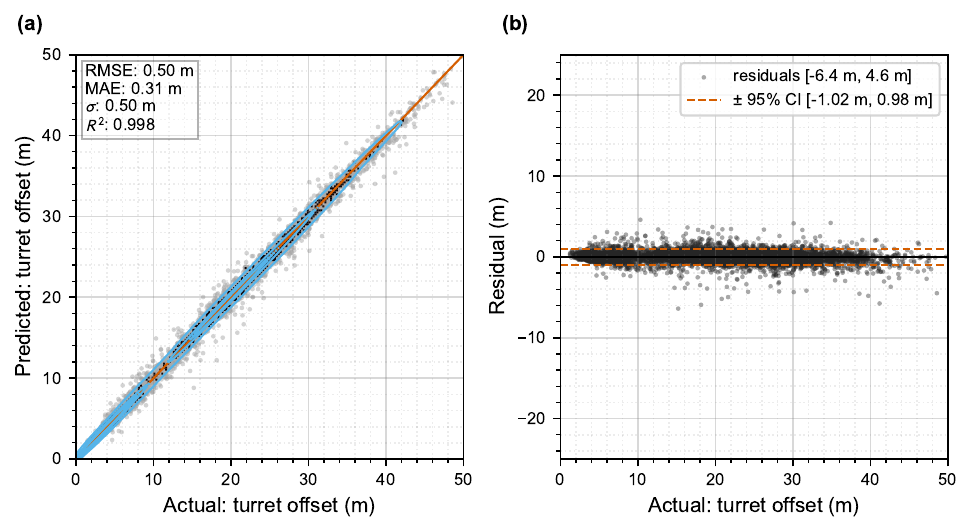}
\caption{\label{fig:turret-offset-XGB-prediction}XGBoost Regression model performance for the most probable maximum turret offset prediction. (a) Actual versus predicted values with the solid red line denoting a perfect fit and blue contours indicating lines of equal probability density. (b) Model residuals \(e_{i}\) with dashed red lines denoting the 95\% bootstrap confidence intervals (CI) for the mean residual.}
\end{figure}

Tables \ref{tab:turret-offset-metrics},
\ref{tab:turret-direction-metrics}, and \ref{tab:tension-metrics} present
performance metrics across the seven machine learning models for
predicting the MPM turret offset magnitude, its corresponding offset
direction, and the MPM omni-directional mooring line fairlead tension,
respectively. Gradient boosting ensemble methods (XGBoost, LightGBM,
and HGB) consistently demonstrated excellent performance across each
response variable. For the turret offset magnitude
(Tab. \ref{tab:turret-offset-metrics}), XGBoost achieved the lowest
errors (\(\RMSE = 0.5049\m\), \(\MAE = 0.3075\m\)) and the highest
coefficient of determination (\(\Rsq = 0.9977\)). In predicting the
direction of the MPM turret offset
(Tab. \ref{tab:turret-direction-metrics}), XGBoost and LightGBM showed
comparable performance (\(\MAE \approx
2.5\degree\text{--}2.9\degree\)). For the fairlead tension
(Tab. \ref{tab:tension-metrics}), XGBoost again demonstrated good
performance with notably lower errors (\(\RMSE = 57.44\kN\), \(\MAE =
29.26\kN\)) when compared to other methods. On the other hand, the
single decision tree consistently demonstrated poor performance. The
neural network (MLP) performed competitively, particularly during
direction prediction, but did not surpass the gradient boosting
methods for other responses. Similarly, the Random Forest model
maintained intermediate performance levels and is on par with the MLP
model. Unsurprisingly, all models significantly outperform linear
regression across all responses and performance metrics.

Using an independent validation dataset,
Figures \ref{fig:turret-offset-XGB-prediction}a and
\ref{fig:turret-offset-XGB-prediction}b demonstrate XGBoost's ensemble
performance through Q-Q and residual plots for the MPM turret offset
magnitude, respectively. The XGBoost model demonstrated significantly
improved prediction accuracy over the linear regression model
(cf. Figures \ref{fig:turret-offset-linear-QQ-prediction}), with an
\(\RMSE\) of \(0.50\m\), \(\MAE=0.31\m\) and \(\Rsq\) of \(0.998\). The
predictions align closely with the actual values across the full range
of MPM turret offsets (\(0\text{--}50\m\)), showing no systematic
bias. The residual plot (Fig. \ref{fig:turret-offset-XGB-prediction}b)
reveals a uniform scatter around zero characterized by a substantially
reduced range (\(-6.4\m\) to \(4.6\m\)) and much narrower 95\% confidence
intervals [\(-1.02\m\), \(0.98\m\)]. These homoscedastic errors indicate
an effective capture of the nonlinear relationships between the
response statistic and its independent variables by the model.

\begin{table*}[!t]
\caption{\label{tab:turret-offset-metrics}Performance metrics for the most probable maximum turret offset (m). Metrics include: Root Mean Squared Error (\(\RMSE\)), Mean Absolute Error (\(\MAE\)), Coefficient of Determination (\(\Rsq\)), Mean residual (\(\bar{e_i}\)), Minimum residual (\(\min(e_i)\)), and Maximum residual (\(\max(e_i)\)).}
\centering
\begin{tabular*}{\textwidth}{@{\hspace{6pt}}l@{\extracolsep{\fill}}cccccc@{\hspace{6pt}}}
\toprule
Model & \(\RMSE\) & \(\MAE\) & \(\Rsq\) & \(\bar{e_i}\) & \(\min(e_i)\) & \(\max(e_i)\)\\
\midrule
Linear Regression & 4.226 & 3.023 & 0.8380 & -0.001175 & -23.26 & 16.57\\
Decision Tree & 1.049 & 0.5728 & 0.9900 & -0.006956 & -10.40 & 13.28\\
Random Forest & 0.7694 & 0.4719 & 0.9946 & 0.007159 & -8.069 & 7.404\\
MLP & 0.6379 & 0.4213 & 0.9963 & 0.1406 & -6.625 & 8.321\\
Histogram Gradient Boosting & 0.5610 & 0.3499 & 0.9971 & -0.001223 & -7.137 & 4.674\\
LightGBM & 0.5293 & 0.3293 & 0.9975 & -0.003366 & -6.482 & 4.710\\
XGBoost & 0.5049 & 0.3075 & 0.9977 & -0.003284 & -6.394 & 4.593\\
\bottomrule
\end{tabular*}
\end{table*}

\begin{table*}[!h]
\caption{\label{tab:turret-direction-metrics}Performance metrics for the direction of the most probable maximum turret offset (\(\mathrm{degrees}\)). Metrics include: Root Mean Squared Error (\(\RMSE\)), Mean Absolute Error (\(\MAE\)), Coefficient of Determination (\(\Rsq\)), Mean residual (\(\bar{e_i}\)), Minimum residual (\(\min(e_i)\)), and Maximum residual (\(\max(e_i)\)).}
\centering
\begin{tabular*}{\textwidth}{@{\hspace{6pt}}l@{\extracolsep{\fill}}cccccc@{\hspace{6pt}}}
\toprule
Model & \(\RMSE\) & \(\MAE\) & \(\Rsq\) & \(\bar{e_i}\) & \(\min(e_i)\) & \(\max(e_i)\)\\
\midrule
Linear Regression & 23.04 & 16.41 & 0.8769 & 1.368 & -175.2 & 170.7\\
Decision Tree & 21.36 & 13.92 & 0.8942 & -0.4628 & -171.8 & 159.1\\
Random Forest & 6.687 & 3.110 & 0.9896 & 0.00542 & -101.2 & 170.1\\
MLP & 5.740 & 2.990 & 0.9924 & -0.1836 & -91.03 & 130.8\\
Histogram Gradient Boosting & 6.261 & 3.008 & 0.9909 & 0.01418 & -172.7 & 147.0\\
LightGBM & 5.846 & 2.854 & 0.9921 & -0.01662 & -162.1 & 142.4\\
XGBoost & 5.925 & 2.493 & 0.9919 & -0.02487 & -169.9 & 163.2\\
\bottomrule
\end{tabular*}
\end{table*}

\begin{table*}[!h]
\caption{\label{tab:tension-metrics}Performance metrics for omni-directional fairlead tension (\(\mathrm{kN}\)). Metrics include: Root Mean Squared Error (\(\RMSE\)), Mean Absolute Error (\(\MAE\)), Coefficient of Determination (\(\Rsq\)), Mean residual (\(\bar{e_i}\)), Minimum residual (\(\min(e_i)\)), and Maximum residual (\(\max(e_i)\)).}
\centering
\begin{tabular*}{\textwidth}{@{\hspace{6pt}}l@{\extracolsep{\fill}}cccccc@{\hspace{6pt}}}
\toprule
Model & \(\RMSE\) & \(\MAE\) & \(\Rsq\) & \(\bar{e_i}\) & \(\min(e_i)\) & \(\max(e_i)\)\\
\midrule
Linear Regression & 308.6 & 196.6 & 0.9117 & -2.078 & -3974. & 1036.\\
Decision Tree & 272.7 & 158.4 & 0.9311 & -2.631 & -4199. & 1657.\\
Random Forest & 68.14 & 39.10 & 0.9957 & -0.9467 & -1428. & 507.5\\
MLP & 68.25 & 45.40 & 0.9957 & -3.281 & -596.8 & 750.4\\
Histogram Gradient Boosting & 66.50 & 36.71 & 0.9959 & 0.1264 & -1067. & 990.6\\
LightGBM & 64.68 & 35.21 & 0.9961 & -0.4804 & -1648. & 1259.\\
XGBoost & 57.44 & 29.26 & 0.9970 & -0.03715 & -1769. & 883.1\\
\bottomrule
\end{tabular*}
\end{table*}
\subsection{Practical application and model deployment}
\label{sec:practical-application}
The ML workflow developed in this study has been deployed in
production, providing forecast and real-time predictions of vessel and
mooring system responses within a much larger facility monitoring
system. In production, the model receives forecast metocean data (over
a 72 hour time window) every 12 hours and real-time metocean data
through a WaveRider buoy Application Programming Interface (API)
request. The model achieves inference times under 30 seconds for
multiple nonlinear response statistics. Achieving comparable accuracy
using a traditional time-domain approach would require at least
30-minutes of sequential CPU wall time to account for multiple
realizations per sea state and to conduct an extreme value analysis.

While the model was originally developed and deployed to support turret
disconnection planning for a single point mooring under onerous
metocean conditions, its ability to forecast could also assist in the
scheduling of side-by-side or tandem offloading
operations and also for the optimization of heading control
strategies. Moreover, the inference times achieved from this model
suggest that it is also suitable for mooring system fatigue
monitoring.

The workflow's generality enables its application to any moored
floating facility, given sufficient training data comprising diverse
metocean conditions and corresponding system response
statistics. While the model's accuracy depends on the fidelity of the
underlying numerical model to represent the as-built vessel and its
mooring system in response to metocean conditions, the approach
presented provides a valuable monitoring tool:

\begin{itemize}
\item when on-board instrumentation is limited or absent
\item when historical measured data are not available
\item during early facility life before sufficient data accumulation.
\end{itemize}

During operation, the model is resource-efficient and can be run on a
single CPU; conducting a heading analysis, computing all critical
responses sequentially and completing all API requests and I/O
operations within the temporal discretization and request time of a
typical WaveRider Buoy (under 60 seconds).

Since the workflow depends on the accuracy of the as-built facility
model, retraining will be required should the characteristics of the
system change (such as weight or mooring line touch-down
locations). Although any regression model can mathematically
extrapolate outside its training domain, this behavior is explicitly
prohibited in production to maintain prediction reliability during
model inference --- in other words, the model is strictly an interpolant and
maps only over its training domain. In terms of maintenance, the model
depends on WaveRider Buoy data and metocean
forecasts, and these require continual monitoring through scheduled test cases to ensure
that any API changes are integrated into the workflow.
\section{Conclusion}
\label{sec:conclusion}
This work demonstrates the effectiveness of machine learning in
predicting nonlinear response statistics over a diverse range of
metocean conditions for a turret-moored floating facility. The
developed model's inference times support both forecasting and
real-time facility monitoring applications. The key findings and
observations from our work include:

\begin{itemize}
\item a successful integration of a physics-based weathervaning solver
within a machine learning workflow
\item gradient boosting methods show excellent performance in predicting
response statistics; a similar performance is anticipated for any
floating facility. In particular: MPM mooring line tensions
predicted to within \(\mathrm{MAE} = 30\kN\), MPM turret offsets to
within \(\mathrm{MAE}=0.31\m\), and directions predicted to within
\(\mathrm{MAE}=2.5\degree\).
\item neural networks (a deep MLP model) can compete, but do not outperform
a tuned gradient boosting method with tabulated data
\item hyperparameter tuning significantly improves model performance, with
a modest 500-trial ensemble showing up to 60\% improvement across all
models considered
\item while this study focused on a generic turret-moored system to
demonstrate broad applicability, the developed workflow has been
successfully deployed on an operational facility and is currently
providing real-time predictions in a production environment.
\end{itemize}

The framework presented here provides a computationally efficient
solution for vessel monitoring and operational
decision-making. Planned future work includes: validating model
performance with field measurements, augmenting the model with field
measurements, incorporating active heading control, extending the
approach to other floating facility types and mooring systems, and
assessing the data generation process to ensure robust domain coverage
and to optimize data sufficiency.
\section*{Acknowledgments}
\label{sec:acknowledgments}
This work was supported by computational resources provided by Worley
Offshore Energy. The authors gratefully acknowledge valuable feedback
from Dmitry Sadovnikov (London Offshore Consultants), Kourosh
Abdolmaleki and Flora Chiew (Worley), as well as the anonymous
reviewers. Publication permission from Worley is greatly
appreciated. The views expressed herein are those of the authors and
do not necessarily represent the views of Worley.

\printbibliography
\end{document}